\documentclass[sigconf]{acmart}
\usepackage{epsfig}
\usepackage{graphicx}
\usepackage{amsmath}
\usepackage{multirow}
\usepackage{subfigure}
\usepackage{balance}

\usepackage{paralist} 

\newenvironment{packed_itemize}{
	\begin{itemize}
		\setlength{\itemsep}{0pt}
		\setlength{\parskip}{0pt}
		\setlength{\parsep}{0pt}
	}{\end{itemize}}

\AtBeginDocument{%
  \providecommand\BibTeX{{%
    \normalfont B\kern-0.5em{\scshape i\kern-0.25em b}\kern-0.8em\TeX}}}

\copyrightyear{2021}
\acmYear{2021}
\setcopyright{acmcopyright}\acmConference[MM '21]{Proceedings of the 29th ACM International Conference on Multimedia}{October 20--24, 2021}{Virtual Event, China}
\acmBooktitle{Proceedings of the 29th ACM International Conference on Multimedia (MM '21), October 20--24, 2021, Virtual Event, China}
\acmPrice{15.00}
\acmDOI{10.1145/3474085.3475197}
\acmISBN{978-1-4503-8651-7/21/10}

\begin{document}
\fancyhead{}
\title{Annotation-Efficient Untrimmed Video Action Recognition}

\author{Yixiong Zou$^{1,3}$, Shanghang Zhang$^2$, Guangyao Chen$^1$} \author{Yonghong Tian$^{1*}$, Kurt Keutzer$^2$, Jos\'e M. F. Moura$^3$}
\thanks{$^*$ indicates corresponding author.}
\affiliation{%
	\institution{Peking University$^1$, University of California, Berkeley$^2$, Carnegie Mellon University$^3$}
	\country{}
}
\email{{zoilsen, gy.chen, yhtian}@pku.edu.cn, {shz, keutzer}@eecs.berkeley.edu, moura@andrew.cmu.edu}

\renewcommand{\shortauthors}{Yixiong Zou, et al.}

\begin{abstract}
	Deep learning has achieved great success in recognizing video actions, but the collection and annotation of training data are still laborious, which mainly lies in two aspects: (1) the amount of required annotated data is large; (2) temporally annotating the location of each action is time-consuming.
	Works such as few-shot learning or untrimmed video recognition have been proposed to handle either one aspect or the other. However, very few existing works can handle both aspects simultaneously.
	In this paper, we target a new problem, \textbf{Annotation-Efficient Video Recognition}, to reduce the requirement of annotations for both large amount of samples and the action locations.
	Challenges of this problem come from two folds: (1) untrimmed videos with only weak supervision; (2)
	video segments not relevant to current actions of interests (background, BG) could contain actions of interests (foreground, FG) in novel classes,
	which widely exists but has rarely been studied in few-shot untrimmed video recognition.
	To achieve this goal, by analyzing the property of BG, we categorize BG into informative BG (IBG) and non-informative BG (NBG),
	and we propose (1) an open-set detection based method to find the NBG and FG, (2) a contrastive learning method for self-supervised learning of IBG and distinguishing NBG, and (3) a self-weighting mechanism for the better distinguishing of IBG and FG.
	Extensive experiments on ActivityNet v1.2 and ActivityNet v1.3 verify the effectiveness of the proposed methods.
\end{abstract}




\maketitle

\section{Introduction}

\begin{figure}[t]
	\subfigure{\includegraphics[width=1.0\columnwidth]{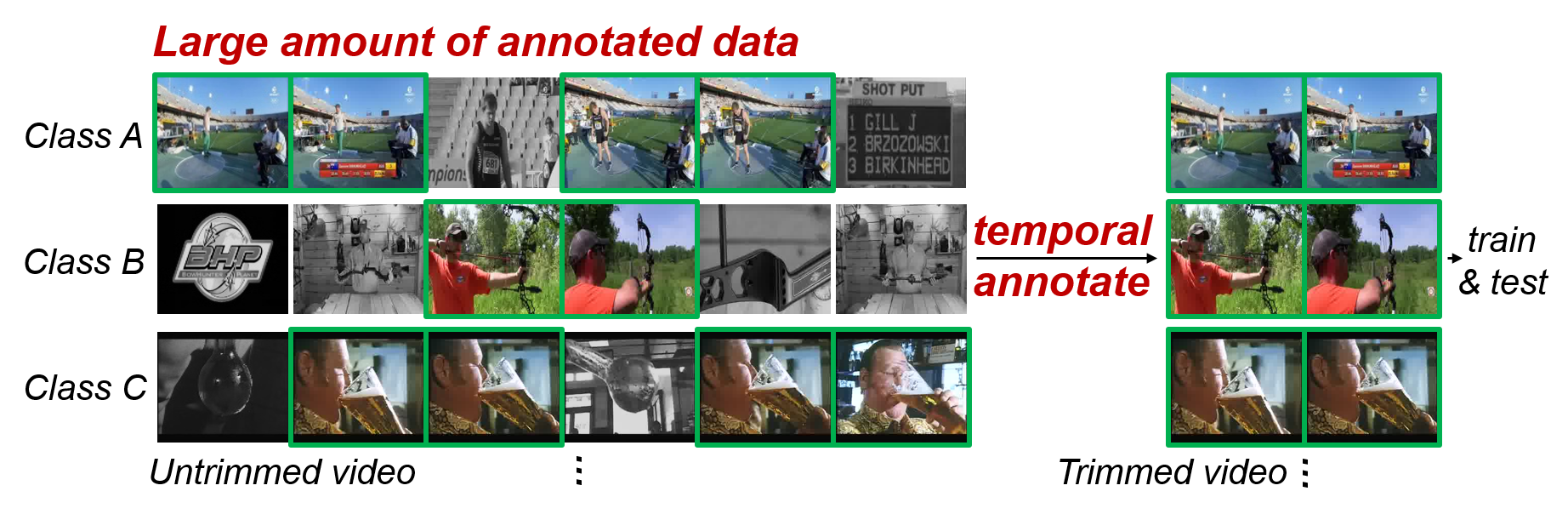}} \\ \vspace{-0.2cm}
	\subfigure{\includegraphics[width=1.0\columnwidth]{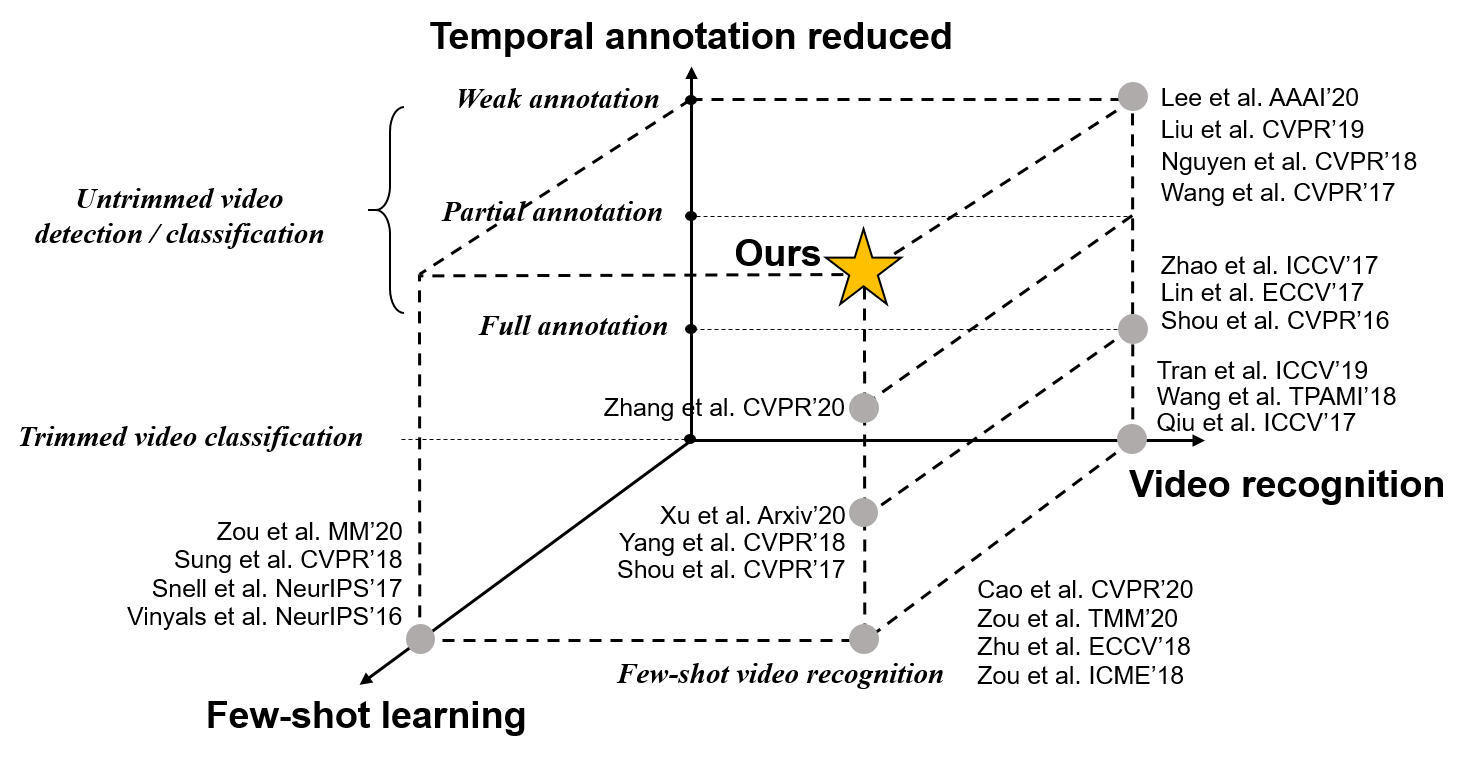}} \vspace{-0.7cm}
	\caption{
		Top: to recognize videos from action class A, B and C, the current data collection and annotation are still laborious, which mainly lies in two aspects: (1) the required amount of labeled data is large and (2) temporally annotating the start and end time (location) of each action is time-consuming.
		Bottom: to handle this problem, works are proposed to alleviate either one aspect or the other (i.e., few-shot learning or reducing temporal annotation). However, reducing both of them simultaneously has rarely been studied, limiting the realistic application of the existing methods. Therefore, we propose the Annotation-Efficient Video Recognition problem (star) to reduce the annotations of both the large amount of data and the action location.} \vspace{-0.2cm}
	\label{fig: motivation}
\end{figure}

Recently, deep learning has achieved great success in video action recognition~\cite{carreira2017quo,kay2017kinetics,wang2018temporal}. However, to recognize videos, the training of deep neural networks still requires large amount of labeled data~\cite{kay2017kinetics,carreira2017quo}, which makes the data collection and annotation laborious in two aspects: (1) the amount of required annotated data is large, and (2) temporally annotating the start \& end time (location) of each action is time-consuming (as shown in Fig.~\ref{fig: motivation} top). What's more, the cost and difficulty of annotating videos is much higher than that of annotating images~\cite{cao2020few}, limiting the realistic applications of existing methods. Therefore, it is of great significance to reduce the requirement of annotations for video action recognition. 

\begin{figure}[t]
	\centering\includegraphics[width=0.8\columnwidth]{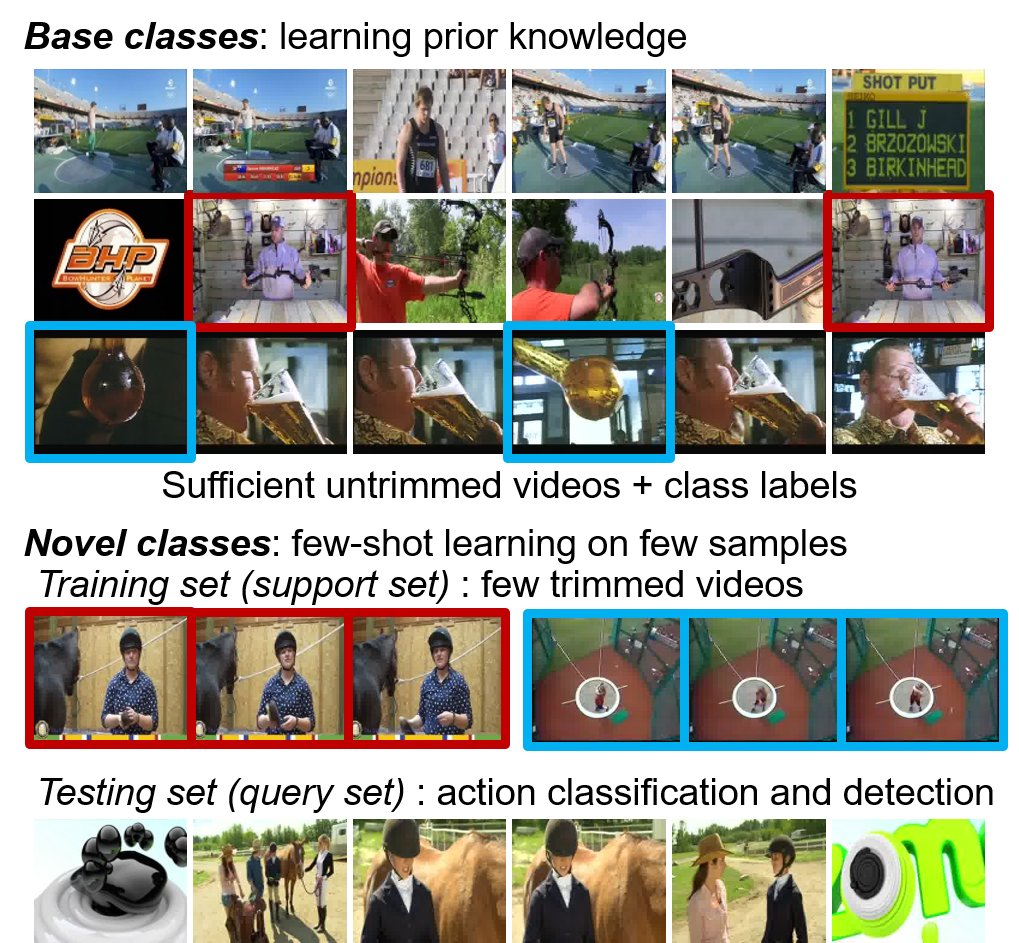} \vspace{-0.3cm}
	\caption{
		Task setting: Following current few-shot learning works~\cite{zhang2020metal, shou2017cdc, yang2018one, xu2020revisiting}, we have two disjoint set of classes, i.e., base classes and novel classes. Our model is first trained on base classes to learn prior knowledge, where only untrimmed videos with class labels are available. Then the model conducts few-shot learning on non-overlapping novel classes with only few trimmed videos, and finally the model is evaluated on novel-class testing videos (untrimmed) by classification and action detection.
		\textit{Challenge}: overlapped base-class background and novel-class foreground, i.e., 
		video segments not relevant to base-class actions could be similar to novel-class actions outlined in the same color
		} \vspace{-0.5cm}
	\label{fig: setting}
\end{figure}

To reduce the amount of annotated samples, few-shot video recognition~\cite{zhu2018compound,zou2018hierarchical,zou2020adaptation,cao2020few} is proposed to recognize novel classes with only a few training samples, with prior knowledge transferred from un-overlapped base classes where sufficient training samples are available. However, most of existing works assume the videos are trimmed in both base classes and novel classes, which still requires temporal annotations to trim videos during data preparation. To reduce annotating action locations, untrimmed video recognition~\cite{wang2017untrimmednets,lee2020background,nguyen2018weakly} has been proposed recently. However, some of the existing works still require temporal annotations of the action location~\cite{shou2016temporal,lin2018bsn,zhao2017temporal}. Others, although can be carried out with only weak supervision (i.e., class label)~\cite{zhao2017temporal,nguyen2019weakly,lee2020background}, are under the traditional close-set setting (i.e., testing classes are the same as training classes), which still requires large amount of labeled samples for the class to recognize.
Combining all above, there is the few-shot untrimmed video recognition problem~\cite{zhang2020metal, shou2017cdc, yang2018one, xu2020revisiting}. However, some of them still require the full temporal annotations for all videos~\cite{shou2017cdc, yang2018one, xu2020revisiting}, and others require large amount of trimmed videos (which we term the partial annotation)~\cite{zhang2020metal}.
As summarized in Fig.~\ref{fig: motivation} bottom, very few works have been done to address all these challenges simultaneously.
Therefore, in this paper, we target to solve a new problem, Annotation-Efficient Video Recognition, where we need to 
recognize novel-class untrimmed testing videos with only few trimmed training videos (i.e., few-shot learning), with prior knowledge transferred from un-overlapped base classes where only untrimmed videos and class labels are available (i.e., weak supervision), as shown in Fig.~\ref{fig: setting}. Note that although on the novel-class training set trimmed videos are required, the annotation cost is limited as only very few samples (e.g, 1-5 samples per novel class) need to be temporally annotated.

The proposed problem has the following two challenges: (1) \textbf{Untrimmed videos with only weak supervision}: videos from both base classes and novel-class testing set are untrimmed (containing non-action video segments, background, BG), and no location annotations are available for distinguishing BG and the video segments with actions (i.e., foreground, FG).
(2) \textbf{Overlapped base-class background and novel-class foreground}: 
BG segments in base classes could be similar to FG in novel classes with similar appearances and motions.
For example, in Fig.~\ref{fig: setting}, frames outlined in red and blue in base classes are BG, but the outlined frames in novel classes are FG, which share similar appearances and motions with the frame outlined in the same color. 
This problem exists because novel classes could contain any kinds of actions not in base classes, including the ignored actions in the base-class background.
If the model learns to force the base-class BG to be away from the base-class FG~\cite{wang2017untrimmednets,lee2020background,nguyen2018weakly}, it will tend to learn non-informative features with suppressed activation (validated in Fig.~\ref{fig: heatmap}) on BG. However, when transferring knowledge to novel-class FG with similar appearances and motions, the extracted features will also tend to be non-informative (validated in Fig.~\ref{fig: heatmap_novel}), harming the novel-class recognition.
To the best of our knowledge, although this difficulty widely exists when transferring knowledge to novel classes, it is rarely studied in few-shot untrimmed video recognition~\cite{zhang2020metal, shou2017cdc, yang2018one, xu2020revisiting}, and this work is the first attempt to handle this problem.

To address the first challenge, a simple way is to develop a method for BG pseudo-labeling~\cite{lee2020uncertainty} or to softly learn to distinguish BG and FG by the attention mechanism~\cite{nguyen2018weakly}. 
To handle the second challenge, we first analyze properties of BG and FG. We find that BG can be coarsely divided into informative BG (IBG) and non-informative BG (NBG). For NBG, there are no informative objects or movements, such as the logo at the beginning of a video (like the left most frame of second row in Fig.~\ref{fig: setting}) or the end credits at the end of a movie, which are not likely to be the cue of recognition. For IBG, there still exist informative objects or movements in video segments, such as the outlined frames in Fig.~\ref{fig: setting}, which is possible to be the FG of novel-class videos, and thus should not be forced to be away from FG during the base-class training.
For NBG, the model should compress its feature space and pull it away from FG, while for IBG, the model should not only capture the semantic objects or movements in it, but also still be able to distinguish it from FG.
Current methods~\cite{wang2017untrimmednets,lee2020background,nguyen2018weakly} simply view these two kinds of BG equivalently and thus tend to harm the novel-class FG features.

Based on the above analysis, we propose our solution to handle these two challenges by viewing these two kinds of BG differently. Our model focuses on the base-class training. 
Firstly, to find NBG, we propose an open-set detection~\cite{hendrycks2016baseline,chen2020learning,dhamija2018reducing} based method for segment pseudo-labeling, which also finds FG and handles the first challenge by pseudo-labeling BG.
Then, a contrastive learning method is introduced for self-supervised learning of informative objects and motions in IBG and distinguishing NBG.
Thirdly, to softly distinguish IBG and FG as well as alleviate the problem of great diversity in the BG class, we propose to softly learn each video segment's attention value by its transformed similarity with the pseudo-labeled BG, which we term as the self-weighting mechanism,
which also handles the first challenge by softly distinguishing BG and FG.
Finally, after base-class training, the prototype-based Nearest Neighbor classification~\cite{snell2017prototypical} and action detection will be performed on novel classes for few-shot recognition.

In all, our contributions can be summarized as follows:
\vspace{-0.05cm}
\begin{packed_itemize}
	\item To reduce the annotations of both the large amount of data and the action location, we define the Annotation-Efficient Video Recognition problem.
	\item To the best of our knowledge, this work is the first attempt to handle the challenge of overlapped base-class BG and novel-class FG in few-shot untrimmed video recognition.
	\item By analyzing the property of BG, we propose (1) an open-set detection based method to find the NBG and FG, (2) a contrastive learning method for self-supervised learning of IBG and distinguishing NBG, and (3) a self-weighting mechanism for the better distinguishing of IBG and FG.
	\item Extensive experiments on ActivityNet v1.2 and ActivityNet v1.3 by both action classification and detection demonstrate the effectiveness and the state-of-the-art performance of the proposed methods.
\end{packed_itemize}

\vspace{-0.2cm}
\section{Related Work}

\subsection{Untrimmed video recognition}
Untrimmed video recognition is proposed to recognize actions in the given untrimmed video~\cite{zhao2017temporal}. 
Typical methods can be grouped into hard-classification based model~\cite{wang2017untrimmednets,lee2020uncertainty} which takes the concept of multiple instance learning (MIL)~\cite{dietterich1997solving} to select video-segments with highest probabilities for classification, and soft-classification based model~\cite{nguyen2018weakly,nguyen2019weakly,min2020adversarial,liu2019completeness}, which learns an attention score for each video segment and use the weighted sum of all segments for classification.
The concept of background modeling has also been applied in \cite{lee2020background,nguyen2019weakly,min2020adversarial,lee2020uncertainty}.
In terms of supervision, these methods can be grouped into full-supervision based methods~\cite{shou2016temporal,lin2018bsn,zhao2017temporal}, which utilize both the class label and the action location to train the model, and weakly-supervised based methods~\cite{zhao2017temporal,nguyen2019weakly,lee2020background} which only have access to the class label.
However, these works are mostly conducted under the close-set setting, and usually large amount of data are needed. In real-world applications, not all classes of actions are easy to be collected and annotated such as the anomaly data~\cite{sultani2018real}.
Therefore, to alleviate the need of large amount of annotated data, we define the Annotation-Efficient Video Recognition problem to consider the weakly-supervised untrimmed video recognition under the few-shot learning setting.

\vspace{-0.2cm}
\subsection{Few-shot video recognition}

Few-shot video recognition is recently proposed to recognize novel-class videos with few training samples~\cite{zhu2018compound,zou2018hierarchical,zou2020adaptation,cao2020few}. For example, \cite{zhu2018compound} designed a memory network to handle this problem. \cite{cao2020few} finds the optimal path along the time axis to compare two videos for better embedding learning.
However, these works assume trimmed videos in both the base classes and novel classes, which is less realistic since trimmed videos need laborious temporal annotations for data preparation. 
Another stream of works is the few-shot untrimmed video recognition~\cite{zhang2020metal, shou2017cdc, yang2018one, xu2020revisiting}, which makes the setting more realistic by considering the untrimmed videos. However, some of them still require the full temporal annotations for all videos~\cite{shou2017cdc, yang2018one, xu2020revisiting}, and others require large amount of trimmed videos (which we term the partial annotation)~\cite{zhang2020metal}.
Also, the phenomenon of overlapped base-class BG and novel-class FG is rarely considered.

\section{Methodology}

\begin{figure}
	\centering
	\includegraphics[width=1.0\columnwidth]{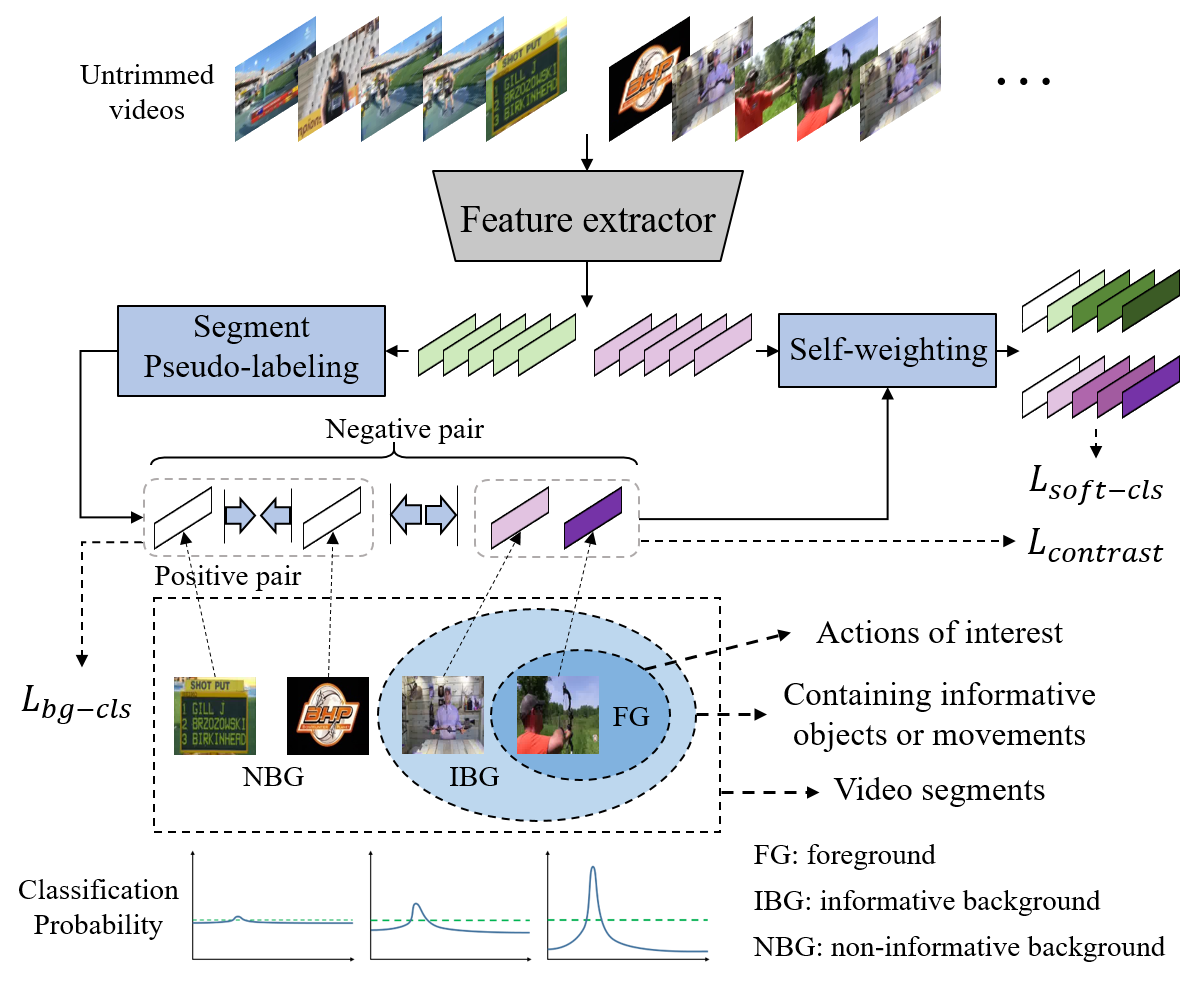} \vspace{-0.8cm}
	\caption{Framework. Our model focuses on the base-class training. Given base-class untrimmed videos, we first find the non-informative background (NBG) segments by each segment's classification probabilities, and pseudo-label segments as NBG by the probabilities closest to the uniform distribution (green dashed line). Similarly, then we pseudo-label segments as informative background (IBG) and foreground (FG). An auxiliary classification ($L_{bg-cls}$) loss is added for NBG modeling, and a self-supervised contrastive loss ($L_{contrast}$) between NBG and IBG + FG is added for capturing informative objects and motions in IBG. BG is also applied in self-weighting each video segment for automatic learning IBG and FG, and a classification loss ($L_{soft-cls}$) is applied for base-class classification.	
	After the base-class training, the prototype-based Nearest Neighbor classification and detection will be performed on novel classes.}
	\label{fig: framework}
\end{figure}

The framework of our method is in Fig.~\ref{fig: framework}. In this section we first give a formal definition of the proposed problem, then analyze its challenges, and finally provide our proposed solutions.

\vspace{-0.2cm}
\subsection{Problem setting}
\label{sec: problem}

To define this problem formally, we follow the current few-shot learning (FSL) problem~\cite{Vinyals2016Matching,snell2017prototypical} to assume there are two disjoint datasets $\mathcal{D}_{base}$ and $\mathcal{D}_{novel}$, with base classes $\mathcal{C}_{base}$ and novel classes $\mathcal{C}_{novel}$ respectively. Note that $\mathcal{C}_{base} \bigcap \mathcal{C}_{novel} = \emptyset$. On $\mathcal{C}_{base}$, sufficient training samples are available, while on $\mathcal{C}_{novel}$, only few training samples are accessible, i.e., few-shot training samples.
As shown in Fig.~\ref{fig: setting}, the model is first trained on $\mathcal{C}_{base}$ for prior knowledge learning, and then the model is trained on the training set (a.k.a support set) of $\mathcal{C}_{novel}$ for the learning with few samples. Finally, the model is evaluated on the testing set (a.k.a query set) of $\mathcal{C}_{novel}$.
For fair comparison, usually there are $K$ classes in the support set and $n$ training samples in each class (a.k.a $K$-way $n$-shot).
Therefore, during the novel-class period, usually numerous $K$-way $n$-shot support set with its query set will be sampled. Each pair of support set and query set can be viewed as an individual small dataset (a.k.a episode) with its training set (i.e., support set) and testing set (i.e., query set) that share the same label space. On novel classes, the sampling-training-evaluating procedure will be repeated on thousands of episodes to obtain the final performance. 

Current FSL works on videos~\cite{zhu2018compound,zou2018hierarchical,zou2020adaptation,cao2020few} assume trimmed videos in both $\mathcal{C}_{base}$ and $\mathcal{C}_{novel}$, which is less realistic due to the laborious temporal annotation of action locations.
Another stream of works, few-shot untrimmed video recognition~\cite{zhang2020metal, shou2017cdc, yang2018one, xu2020revisiting}, although can be carried on untrimmed videos under the FSL setting, still requires either the full temporal annotation~\cite{shou2017cdc, yang2018one, xu2020revisiting} or the partial temporal annotation~\cite{zhang2020metal} (i.e., large amount of trimmed videos) on base classes for distinguishing the action part (foreground, FG) and non-action part (background, BG) of video. As base classes require large amount of data for prior knowledge learning, it is still costly for data preparation.
To solve this problem, we proposed the Annotation-Efficient Video Recognition problem, where in $\mathcal{C}_{base}$ only untrimmed videos with class labels (i.e., weak supervision) are available, and in $\mathcal{C}_{novel}$ only few trimmed videos are used for support set while untrimmed videos are used for query set for action classification and detection.
Note that although trimmed videos are needed for the support set, the cost of temporal annotation is limited since only few samples need to be temporally annotated.

Compared with current works, the challenges are in two aspects:

\noindent(1) \textbf{Untrimmed video with only weak supervision}, different with few-shot video recognition~\cite{zhu2018compound,zou2018hierarchical,zou2020adaptation,cao2020few}, which means noisy parts of video (background, BG) exist in both base and novel classes; also different with the full supervision setting~\cite{shou2016temporal,lin2018bsn,zhao2017temporal} or the partial supervision setting~\cite{zhang2020metal}, which means only the class label and untrimmed videos are available in base classes.

\noindent(2) \textbf{Overlapped base-class background and novel-class foreground}, which means BG segments in base classes could be similar to FG in novel classes with similar semantic meaning. 
For example, in Fig.~\ref{fig: setting}, the outlined frames outlined in base classes are BG, but the outlined frames novel classes are FG, which share similar appearances or motions with the frame outlined in the same color. 
To the best of our knowledge, although this difficulty widely exists when transferring knowledge to few-shot novel classes, it is rarely studied in the domain of few-shot untrimmed video recognition~\cite{zhang2020metal, shou2017cdc, yang2018one, xu2020revisiting}, and this work is the first attempt to handle this problem.

In the following sections, we will elaborate our analysis and solutions to these challenges, as shown in Fig.~\ref{fig: framework}.

\vspace{-0.1cm}
\subsection{Baseline model}
\label{sec: baseline}

For better understanding, we first provide a baseline model based on baselines of FSL and untrimmed video recognition, then we give analysis and propose our modifications to this model.
For FSL, a widely adopted baseline model~\cite{li2019large, DBLP:journals/corr/abs-1904-04232, hariharan2017low, qiao2017few, rusu2019meta, tokmakov2019learning,zou2020compositional} is to first classify each base-class video $x$ into all base classes $\mathcal{C}_{base}$, then use the trained backbone network for feature extraction, and finally conduct the Nearest Neighbor classification on novel classes based on the support set and query set. The base-class classification loss: 

\vspace{-0.3cm}
\begin{equation}
L_{cls} = - \sum_{i=1}^{N} y_i log(\frac{e^{\tau {W_i} F(x)}}{\sum_{k=1}^{N} e^{\tau {W_k} F(x)} })
\label{eq:classification loss}
\end{equation}
where $y_i$ is set to 1 if $x$ has the $i$th action, otherwise 0, $F(x) \in R^{d \times 1}$ is the extracted video feature, $d$ is the number of channels, $\tau$ is the temperature parameter and is set to 10.0, $N$ is the number of base classes, and $W \in R^{N \times d}$ is the parameter of the fully-connected (FC) layer for base-class classification (with the bias term abandoned~\cite{DBLP:journals/corr/abs-1904-04232,zou2020compositional}). Note that $F(x)$ is $L_2$ normalized along columns and $W$ is $L_2$ normalized along rows.
The novel-class classification is based on

\vspace{-0.3cm}
\begin{equation}
\begin{aligned}
\hat{y_q} = \{y_i | P(y_i|x^U_q) > t_a\} = 
\{i | \frac{e^{s(F(x^U_q), p^U_i)}}{\sum_{k=1}^{K} e^{s(F(x^U_q), p^U_k)} } > t_a\}
\end{aligned}
\label{eq:MN_P}
\end{equation}
where $x^U_q$ is the novel-class query sample to classify, $\hat{y_q}$ is its predicted label(s), $t_a$ denotes the action threshold, $s(,)$ denotes the similarity function (e.g., cosine similarity), $K$ is the number of classes in the support set, and $p^U_i$ is the prototype for each class. Typically the prototype is calculated as $p^U_i = \frac{1}{n} \sum_{j=1}^{n} F(x^U_{ij})$~\cite{snell2017prototypical} where $x^U_{ij}$ is the $j$th sample in the $i$th class of the support set, and $n$ is the number of sample in each class.

For untrimmed video recognition, to obtain the video feature $F(x)$ given $x$, we follow current works~\cite{wang2017untrimmednets,lee2020background,nguyen2018weakly} to split each video into $T$ un-overlapped segments, where each segment contains $t$ consecutive frames, thus the video can be represented as $x = \{s_i\}_{i=1}^T$ where $s_i$ is the $i$th segment.
As BG exists in $x$, segments contribute unequally to the video feature.
Typically, one widely used baseline~\cite{wang2017untrimmednets,nguyen2018weakly} is the attention-based model, which learns a weight for each segment by a small network, and uses the weighted combination of all segment features as the video feature as
\vspace{-0.1cm}
\begin{equation}
F(x) = \sum_{i=1}^{T} \frac{h(s_i)} {\sum_{k=1}^{T} h(s_k)} f(s_i)
\label{eq:f_soft}
\end{equation}
where $f(s_i) \in R^{d \times 1}$ is the segment feature, which could be extracted by a 3D convolutional neural network, and $h(s_i)$ is the weight for $s_i$.
We denote the above baseline as the soft-classification baseline, and below we will illustrate our analysis and modification to it.

\subsection{Handle Challenges}

To address the challenge of untrimmed videos with weak supervision, a simple way is to develop a method for BG pseudo-labeling~\cite{lee2020uncertainty} or to softly learn to distinguish BG and FG by the attention mechanism~\cite{nguyen2018weakly}. 
To handle the challenge of overlapped base-class BG and novel-class FG, we first analyze properties of BG and FG.

Firstly, BG does not contain the action of interest, which means by removing these parts of video segments, the remaining parts (i.e., FG) could still be recognized as the action of interest.
Therefore, current methods either only utilize the FG in classification~\cite{wang2017untrimmednets,lee2020uncertainty} or softly learn large weights for FG segments and learn small weights for BG segments~\cite{nguyen2018weakly,nguyen2019weakly,min2020adversarial,liu2019completeness}, which makes the supervision from class labels less effective for the model to capture the objects or movements in BG segments (as validated in Fig.~\ref{fig: heatmap}).

Secondly, BG shows great diversity, which means any videos, as long as they are not relevant to the current action of interest, could be recognized as BG. 
However, novel classes could also contain any kinds of actions not in base classes, including the ignored actions in the base-class BG, as shown in Fig.~\ref{fig: setting}. As studied in \cite{zhou2016learning}, deep networks tend to have similar activation given input with similar appearances. If novel-class FG is similar to base-class BG, the deep network might fail to capture semantic objects or movements, as it does on base classes (feature collapse, validated in Fig.~\ref{fig: heatmap_novel}).

However, in the infinite space of BG, empirically not all video segments could be recognized as FG. For example, in the domain of human action recognition~\cite{wang2018temporal,caba2015activitynet,carreira2017quo}, only videos with human and actions could be recognized as FG. In the meanwhile, video segments that provide no information about human are less likely to be recognized as FG in the vast majority of classes, such as the logo page at the beginning of a video, or the end credits at the end of a movie, as shown in Fig.~\ref{fig: setting}\footnote{More examples could be found in the supplementary materials.}.
Therefore, we categorize the BG containing informative objects or movements as the \textit{informative background (IBG)}, and categorize the BG containing less information as the \textit{non-informative background (NBG)}. For NBG, separating it with FG will be less likely to prevent the model from capturing semantic objects or movements in novel-class FG, while for IBG, forcing it to be away from FG would cause such problem. 
Therefore, we find it important to view differently for these two kind of BG, and methods should be specifically developed for them.
\textit{For NBG, the model should compress its feature space and pull it away from FG, while for IBG, the model should not only capture the semantic objects or movements in it, but also still be able to distinguish it from FG.}

Based on the above analysis, below we propose our solution to these challenges. As shown in Fig.~\ref{fig: framework}, our model can be summarized as (1) finding NBG, (2) self-supervised learning of IBG and distinguishing NBG, and (3) automatic learning of IBG and FG.

\vspace{-0.1cm}
\subsubsection{\textbf{Finding NBG}} 
\label{sec: finding}

As defined above, the NBG seldom share semantic objects and movements with FG. Therefore, empirically its feature would be much more distant from FG than the IBG, with its classification probability being much more closer to the uniform distribution (validated in Fig.~\ref{fig: max_logits}).
Considering that BG segments also cannot be classified into any given base classes, we find such criteria is similar to one typical solution~\cite{hendrycks2016baseline} of the open-set detection problem, which rejects unknown samples that do not belong to any base classes~\cite{hendrycks2016baseline,chen2020learning,dhamija2018reducing} by the classification probability being close to the uniform distribution. This inspires us to propose our solution of finding NBG based on open-set detection.
Specifically, given an input untrimmed $x = \{s_i\}_{i=1}^T$ and $N$ base classes, we seek for BG by each segment's maximum classification probability as
\vspace{-0.1cm}
\begin{equation}
i_{bg} = argmin \max P(s_k)
\label{eq:least_certainty}
\end{equation}
where $i_{bg}$ is the index of the BG segment, $P(s_k) \in R^{N \times 1}$ is the base-class logit\footnote{Empirically we find the unnormalized logit is better than the probability distribution.}, calculated as $W f(s_k)$ and $f(s_k)$ is also $L_2$ normalized. For simplicity, we denote the pseudo-labeled BG segment $s_{i_{bg}}$ as $s_{bg}$. Then, we pseudo-label NBG by filtering its max logit as 
\vspace{-0.1cm}
\begin{equation}
\{s_{nb}\} = \{s_{bg} | \max P(s_{bg}) < t_n \}
\label{eq:bg2nbg}
\end{equation}
where $s_{nb}$ denotes the pseudo-labeled NBG, and $t_n$ is the threshold.

In the domain of open-set detection, the pseudo-labeled segment can be viewed as the known-unknown sample~\cite{walter2013toward}, for which another auxiliary class can be added to classify it~\cite{bendale2016towards}, which is consistent with current works that classify soft weighted segments into an auxiliary BG class~\cite{lee2020background,nguyen2019weakly}.
Therefore, we applied a loss for the NBG classification as
\vspace{-0.1cm}
\begin{equation}
\begin{aligned}
& L_{bg-cls} = - log(P(y_{nb}|s_{nb}))  = - log(\frac{e^{\tau {W^E_{nb}} f(s_{nb})}}{\sum_{i=1}^{N} e^{\tau {W^E_i} f(s_{nb})}} )
\label{eq:bg-cls}
\end{aligned}
\end{equation}
where $W^E \in R^{(N+1) \times d}$ denotes the FC parameters expanded from $W$ to include the NBG class, $y_{nb}$ is the label of the NBG.
The most similar idea to us is \cite{lee2020uncertainty}, which pseudo-label BG segments by the feature norm and is conducted in the close-set many-shot setting. However, it pseudo-labels BG by the feature norm and the feature in FSL is always $L_2$ normalized as in section~\ref{sec: baseline}, therefore it could not be applied in our problem.
Also, by finding the NBG, we are also solving the first challenge (untrimmed video with weak supervision) by pseudo-labeling segments as BG.


\vspace{-0.2cm}
\subsubsection{\textbf{Self-supervised learning of IBG and distinguishing \\ NBG}}
\label{sec: contrastive learning}

As analyzed, FG is informative of current actions of interest, containing informative objects and movements, IBG is not informative of current actions of interest but contains informative objects and movements, while NBG is neither informative of current actions nor containing informative objects or movements. The correlation between these three terms is shown in Fig.~\ref{fig: framework}.
As the supervision from class labels could mainly help distinguishing whether one video segment is informative of recognizing current actions, the learning of IBG could not merely rely on the classification supervision because IBG is not informative enough of that task. Therefore, other supervisions are needed for the learning of IBG.

As analyzed, to solve the problem of overlapped base-class BG and novel-class FG, the model need to capture the informative things in IBG, which is just the difference between NBG and IBG + FG.
These inspire us to develop a contrastive learning method by enlarging the distance between NBG and IBG + FG.

Currently, contrastive learning has achieved great success in self-supervised learning, which aims at learning embedding from unsupervised data by constructing positive and negative pairs~\cite{oord2018representation,tian2019contrastive}. 
The distances within positive pairs are reduced, while the distances within negative pairs are enlarged.

In view that in section~\ref{sec: finding} the maximum classification probability also measures the confidence that the given segment belonging to the base classes, and FG always shows the highest confidence~\cite{wang2017untrimmednets,sultani2018real}, we also utilize such criteria for pseudo-labeling FG, which is symmetric to the BG pseudo-labeling and consistent with MIL~\cite{wang2017untrimmednets}.
Compared with current works, we not only pseudo-label segments with highest confidence as the FG segments, \textit{but also include some segments with relatively high confidence as the pseudo-labeled IBG}. The insight is that since IBG shares informative objects or movements with FG, its action score should be smoothly decreased from FG, therefore the confidence score between FG and IBG could be close (validated in Fig.~\ref{fig: max_logits}). However, it is hard to set a threshold for distinguishing FG and IBG, but we are not aiming to distinguishing them in this loss (specifically, the distinguishing is in section~\ref{sec: automatic}), therefore, we could simply choose segments with top confidences (but the number of chosen segments is larger than MIL) to be the pseudo-labeled FG and IBG, and mark features from NBG and FG + IBG as the negative pair, for which we need to enlarge the distance.
The capturing of informative appearances is validated in Fig.~\ref{fig: heatmap} and Fig.~\ref{fig: heatmap_novel}.

For the positive pair, since we need to compress the feature space of NBG,
we mark two NBG features as the positive pair, for which we need to reduce the distance.
Note that we cannot set features from the FG and IBG as the positive pair, because IBG does not help the base-class recognition, thus such pairs would harm the model.

Specifically, given a batch of untrimmed videos with batch size $B$, we take all NBG segments $\{s^j_{bg}\}_{j=1}^B$ and FG + IBG segments $\{s^j_{fg+ibg}\}_{j=1}^B$ to calculate the contrastive loss as 

\vspace{-0.2cm}
\begin{equation}
\begin{aligned}
&L_{contrast} =  \max_{j \neq k} d(f(s^j_{nb}), f(s^k_{nb})) \\
&				+ \beta \max(0, margin - \min d(f(s^j_{fg+ibg}), f(s^k_{nb})))
\label{eq:contrast}
\end{aligned}
\end{equation}
where $d(,)$ denotes the squared Euclidean distance between two $L_2$ normalized vectors, and $margin$ is set to 2.0.

\subsubsection{\textbf{Automatic learning of IBG and FG}}
\label{sec: automatic}

For IBG, we cannot explicitly force its separation with FG, but the model should still be able to distinguish it from FG.
To achieve this goal, we look back into our attention-based baseline model, which automatically learns to distinguish BG and FG by learning a weight for each segment via a global weighting network.
However, this model possibly has one drawback in our setting of problem: it assumes a global weighting network for the BG class, which implicitly assumes a global representation of the BG class. However, the BG class always shows great diversity,
which is even exaggerated when transferring the model to un-overlapped novel classes, since greater diversity not included in the base classes could be introduced in novel classes now.
This drawbacks hinder the automatic learning of IBG and FG, which inspires us to propose our solution for alleviating it.

Our solution is to abandon the assumption about the global representation of BG. Instead, for each untrimmed video, we propose to use its pseudo-labeled BG segment to measure the importance of each video segment, and use its transformed similarity to be the attention value, which we term the \textit{self-weighting} mechanism.

Specifically, we denote the pseudo-labeled BG segment for video $x = \{s_i\}_{i=1}^T$ as $s_{bg}$ as in Eq.~\ref{eq:least_certainty}. Since the feature extracted by the backbone network is $L_2$ normalized, the cosine similarity between $s_{bg}$ and the $k$th segment $s_k$ can be calculated as $f(s_{bg})^\top f(s_k)$. Therefore, we seek to design a transformation function based on $f(s_{bg})^\top f(s_k)$ to replace the weighting function $h()$ in Eq.~\ref{eq:f_soft}, i.e., $ h(s_k) = g(f(s_{bg})^\top f(s_k))$.
Specifically, the function is defined as 
\vspace{-0.1cm}
\begin{equation}
\begin{aligned}
g(f(s_{bg})^\top f(s_k)) = \frac{1}{1 + e^{- \tau_s (1 - c - f(s_{bg})^\top f(s_k))}}
\label{eq:self-weighting}
\end{aligned}
\end{equation}
where $\tau_s$ controls the peakedness of the score and is set to 8.0, and $c$ controls the center of the cosine similarity which is set to 0.5. We design such function because the cosine similarity between $f(s_{bg})$ and $f(s_k)$ is in range [-1, 1]. In order to map the similarity to [0, 1], we follow \cite{nguyen2018weakly} to add a sigmoid function, and add $\tau_s$ to ensure the max and min weight are close to 0 and 1. In view that two irrelevant vectors should have cosine similarity at 0, we set the center $c$ to 0.5.
Note that this mechanism is different from the self-attention mechanism~\cite{wang2017untrimmednets,nguyen2018weakly}, which uses an extra global network to learn the segment weight from the segment feature itself. Here the segment weight is the transformed similarity with the pseudo-labeled BG, and there is no extra global parameters for the weighting.
The modification of classification in Eq.~\ref{eq:classification loss} is
\vspace{-0.1cm}
\begin{equation}
L_{cls-soft} = -log(\frac{e^{\tau {W^E_y} F(x)}}{\sum_{i=1}^{N+1} e^{\tau {W^E_i} F(x)} })
\label{eq:cls-soft}
\end{equation}
where $W^E \in R^{(N+1) \times d}$ is the FC parameters expanded to include the BG class as in Eq.~\ref{eq:bg-cls}, and $F(x)$ in Eq.~\ref{eq:f_soft} is modified as

\vspace{-0.2cm}
\begin{equation}
F(x) = \sum_{i=1}^{T} \frac{g(f(s_{bg})^\top f(s_i))} {\sum_{k=1}^{T} g(f(s_{bg})^\top f(s_k))} f(s_i)
\label{eq:f-soft-self}
\end{equation}
By such weighting mechanism, we are also solving the first challenge (untrimmed video with weak supervision) by softly learning to distinguish BG and FG.
Combining all above, the model is trained with
\vspace{-0.1cm}
\begin{equation}
L = L_{cls-soft} + \gamma_1 L_{contrast} + \gamma_2 L_{bg-cls}
\label{eq:loss-all}
\end{equation}
where $\gamma_1$ and $\gamma_2$ are hyper-parameters. 
With the proposed methods, as shown in Fig.~\ref{fig: heatmap} and Fig.~\ref{fig: heatmap_novel}, our model is capable of capturing informative objects and movements in IBG, and is still able to distinguish BG and FG, therefore helping the recognition.

\vspace{-0.2cm}
\subsection{Novel-class testing}

After base-class training, on novel classes we first extract the support-set and query features with the trained backbone $f()$. 
For the support set containing trimmed videos, we directly average each segment feature to be the video feature.
For the query set containing untrimmed videos, we pseudo-label the BG segments by the $K$-way logit and all segments will be weighted averaged as above to obtain the feature $F()$ for such video. 
For action classification, evaluation will be conducted as Eq.\ref{eq:MN_P}.
For action detection, the temporal class activation map~\cite{nguyen2018weakly} will be calculated based on the attention score and the support set feature.

\begin{table*}[t]
	\begin{center}
		\caption{Comparison with current works by action detection.} \vspace{-0.3cm}
		\label{tab:sota_all_detection}
		\resizebox{1.9\columnwidth}{!}{
			\begin{tabular}{c|c|c|c|c|c|c|c|c|c}
				\hline\hline
				\multirow{3}{*}{Method}& \multirow{3}{*}{Supervision} & \multicolumn{4}{c|}{ActivityNet v1.2} & \multicolumn{4}{c}{ActivityNet v1.3} \\
				\cline{3-10}
				& & \multicolumn{2}{c|}{mAP@0.5} & \multicolumn{2}{c|}{average mAP} & \multicolumn{2}{c|}{mAP@0.5} & \multicolumn{2}{c}{average mAP} \\ \cline{3-10}
				& & 5-way 1-shot & 5-way 5-shot & 5-way 1-shot & 5-way 5-shot & 5-way 1-shot & 5-way 5-shot & 5-way 1-shot & 5-way 5-shot \\ \hline

				CDC~\cite{shou2017cdc} & Full & $ 8.2 $ & $ 8.6 $ & $ 2.4 $ & $ 2.5 $ & - & - & - & - \\ 
				Yang~\cite{yang2018one} & Full & $ 22.3 $ & $ 23.1 $ & $ 9.8 $ & $ 10.0 $ & - & - & - & - \\ 
				F-PAD-ctrl~\cite{xu2020revisiting} & Full & $ \textbf{31.7} $ & $ 37.9 $ & $ \textbf{19.4} $ & $ 23.5 $ & $31.4$ & $39.0$ & $20.8$ & $24.1$ \\
				\hline
				Ours 																					& Weak & $30.24 \pm 0.23$ & $\textbf{38.23} \pm \textbf{0.24}$ & $18.82 \pm 0.16$ & $\textbf{23.81} \pm \textbf{0.17}$ & $\textbf{35.41} \pm \textbf{0.25}$ & $\textbf{43.72} \pm \textbf{0.25}$ & $\textbf{23.20} \pm \textbf{0.19}$ & $\textbf{28.72} \pm \textbf{0.19}$ \\ 
				\hline\hline
		\end{tabular}}
	\end{center}\vspace{-0.2cm}
\end{table*}

\begin{table}[t]
	\begin{center}
		\caption{Comparison with current works by classification.} \vspace{-0.3cm}
		\label{tab:sota_all}
		\resizebox{1.05\columnwidth}{!}{
			\hskip-0.6cm
			\begin{tabular}{c|c|c|c|c}
				\hline\hline
				\multirow{2}{*}{Method} & \multicolumn{2}{c|}{ActivityNet v1.2} & \multicolumn{2}{c}{ActivityNet v1.3} \\ \cline{2-5}
				& 5-way 1-shot & 5-way 5-shot & 5-way 1-shot & 5-way 5-shot \\ \hline
				ImageNet pretraining 	                                                                & $57.62 \pm 0.26$ & $80.86 \pm 0.20$ & $61.77 \pm 0.26$ & $84.47 \pm 0.18$ \\
				MIL~\cite{wang2017untrimmednets} + BL~\cite{DBLP:journals/corr/abs-1904-04232} 	& $65.21 \pm 0.25$ & $80.69 \pm 0.19$ & $70.34 \pm 0.24$ & $85.14 \pm 0.18$ \\ 
				TSN~\cite{wang2018temporal} + BL~\cite{DBLP:journals/corr/abs-1904-04232}		& $65.85 \pm 0.25$ & $81.28 \pm 0.20$ & $70.56 \pm 0.23$ & $85.58 \pm 0.18$ \\
				TCAM~\cite{nguyen2018weakly} + BL~\cite{DBLP:journals/corr/abs-1904-04232} 		& $66.21 \pm 0.27$ & $81.02 \pm 0.19$ & $70.87 \pm 0.26$ & $85.60 \pm 0.18$ \\
				WBG~\cite{nguyen2019weakly} + BL~\cite{DBLP:journals/corr/abs-1904-04232} 		& $66.31 \pm 0.24$ & $81.28 \pm 0.19$ & $71.18 \pm 0.24$ & $85.91 \pm 0.17$ \\
				TSN~\cite{wang2018temporal} + MN~\cite{Vinyals2016Matching} 		            & $66.23 \pm 0.24$ & $81.22 \pm 0.20$ & $71.05 \pm 0.24$ & $85.11 \pm 0.19$ \\
				TCAM~\cite{nguyen2018weakly} + MN~\cite{Vinyals2016Matching} 		            & $66.18 \pm 0.24$ & $81.59 \pm 0.17$ & $70.97 \pm 0.24$ & $85.70 \pm 0.17$ \\
				WBG~\cite{nguyen2019weakly} + MN~\cite{Vinyals2016Matching} 		            & $66.16 \pm 0.25$ & $81.35 \pm 0.19$ & $71.18 \pm 0.24$ & $85.76 \pm 0.17$ \\
				MIL~\cite{wang2017untrimmednets} + MN~\cite{Vinyals2016Matching}	            & $66.73 \pm 0.26$ & $81.85 \pm 0.19$ & $71.78 \pm 0.24$ & $86.01 \pm 0.17$ \\ 
				\hline
				Ours 				& $\textbf{68.53} \pm \textbf{0.22}$ & $\textbf{84.54} \pm \textbf{0.19}$ & $\textbf{73.39} \pm \textbf{0.25}$ & $\textbf{88.06} \pm \textbf{0.17}$ \\ 
				\hline\hline
		\end{tabular}}
	\end{center}\vspace{-0.3cm}
\end{table}

\begin{table}[t]
	\begin{center}
		\caption{Ablation study. Soft: soft-classification baseline; BG: background pseudo-labeling; SW: self-weighting; CL: contrastive learning.} \vspace{-0.3cm}
		\label{tab:ablation_all}
		\resizebox{1.05\columnwidth}{!}{
			\hskip-0.6cm
			\begin{tabular}{cccc|c|c|c|c}
				\hline\hline
				\multicolumn{4}{c|}{Module} & \multicolumn{2}{c|}{ActivityNet v1.2} & \multicolumn{2}{c}{ActivityNet v1.3} \\ \hline
				Soft& BG & SW & CL & 5-way 1-shot & 5-way 5-shot & 5-way 1-shot & 5-way 5-shot \\ \hline
				$\checkmark$ & 				& 				& 				& $65.38 \pm 0.26$ & $81.04 \pm 0.19$ & $70.56 \pm 0.24$ & $85.09 \pm 0.18$ \\
				$\checkmark$ & $\checkmark$ & 				& 				& $66.02 \pm 0.25$ & $82.02 \pm 0.18$ & $71.38 \pm 0.25$ & $86.06 \pm 0.17$ \\
				$\checkmark$ & 				& $\checkmark$ 	& 				& $66.42 \pm 0.24$ & $81.57 \pm 0.20$ & $71.47 \pm 0.23$ & $85.77 \pm 0.18$ \\
				$\checkmark$ & 				& 				& $\checkmark$ 	& $67.33 \pm 0.23$ & $83.42 \pm 0.18$ & $72.26 \pm 0.25$ & $86.97 \pm 0.17$ \\
				$\checkmark$ & $\checkmark$ & $\checkmark$ 	& 				& $67.01 \pm 0.26$ & $82.26 \pm 0.19$ & $71.92 \pm 0.24$ & $86.53 \pm 0.19$ \\
				$\checkmark$ & $\checkmark$ & 				& $\checkmark$ 	& $67.80 \pm 0.26$ & $83.91 \pm 0.18$ & $72.79 \pm 0.22$ & $87.58 \pm 0.16$ \\
				$\checkmark$ & 				& $\checkmark$ 	& $\checkmark$ 	& $67.99 \pm 0.23$ & $83.94 \pm 0.18$ & $72.90 \pm 0.25$ & $87.45 \pm 0.17$ \\
				$\checkmark$ & $\checkmark$ & $\checkmark$ 	& $\checkmark$ 	& $68.53 \pm 0.22$ & $84.54 \pm 0.19$ & $73.39 \pm 0.25$ & $88.06 \pm 0.17$ \\
				\hline\hline
		\end{tabular}}
	\end{center}\vspace{-0.4cm}
\end{table}

\section{Experiments}

To verify the proposed methods on the proposed problem, we conduct experiments on both ActivityNet v1.2 and v1.3~\cite{caba2015activitynet}. We first introduce the datasets and implementation details. Then we compare our method with state-of-the-art and show the ablation study of each module. Due to the space limitation, please refer to the supplementary material for more details.

\vspace{-0.2cm}
\subsection{Datasets and settings}

Experiments are conducted on ActivityNet v1.2 and v1.3~\cite{caba2015activitynet}. ActivityNet v1.2 is originally proposed for close-set untrimmed video recognition, which contains 100 action classes. From its website we can get the video and the annotation of its original training set (4819 videos) and the validation set (2383 videos). We follow \cite{zhang2020metal,yang2018one} to use all 7202 videos and randomly choose 80 classes for base classes and 20 classes for novel classes. 
ActivityNet v1.3 is an extension of ActivityNet v1.2, which consists of 200 action classes and 14950 videos. Similar to v1.2, we use all data from its original training and validation set, and randomly choose 160 classes as base classes and 40 classes as novel classes.
For evaluation, we sample $K$-way $n$-shot episodes as stated in section~\ref{sec: problem}. For classification-based evaluation, the $K$-way classification accuracy together with the 95\% confidence interval will be reported. For detection-based evaluation, the $K$-way mean average precision with tIoU threshold set to 0.5 and ranging from [0.5, 0.95] with 0.05 as the interval will be reported.

\vspace{-0.2cm}
\subsection{Implementation details}
For each untrimmed video, we extract its RGB frames at 25 FPS and at the resolution of 256 $\times$ 256. 
We averagely divide each video into 100 non-overlapping segments and sample 8 frames for each segment (i.e., $T$=100, $t$=8 in section~\ref{sec: baseline}).
The feature is extracted by ResNet50~\cite{He_2016_CVPR}. In view that the widely used Kinetics dataset~\cite{Kay2017The} may share similar action classes with the novel classes~\cite{xu2020revisiting}, we did not follow \cite{zhang2020metal} to use it for pre-training. Instead, we only use ImageNet~\cite{deng2009imagenet} for ResNet50 pre-training. After feature extraction, the ResNet50 and the extracted features are fixed. Then, a spatial transforming layer with spatial kernel size $1 \times 1$ is added to transform the feature into 2048 channels, and a depth-wise temporal convolution layer with kernel of shape $1 \times 8$ is added to capture the temporal information.
We follow \cite{zhang2020metal} to only use the RGB steam.
The model is implemented with TensorFlow~\cite{abadi2016tensorflow}, trained with learning rate at 0.01, and optimized with the Nesterov Momentum Optimizer~\cite{sutskever2013importance}.
For other details, please refer to the supplementary material. 

\vspace{-0.2cm}
\subsection{Comparison with state-of-the-art}


The comparisons with current works are listed in Tab.~\ref{tab:sota_all} and Tab.~\ref{tab:sota_all_detection}. We choose MatchingNet~\cite{Vinyals2016Matching} (MN) and Baseline++~\cite{DBLP:journals/corr/abs-1904-04232} (BL) as the state-of-the-art methods for FSL, choose TSN~\cite{wang2018temporal} with only the RGB stream as the state-of-the-art for video recognition, and choose MIL~\cite{wang2017untrimmednets}, TCAM~\cite{nguyen2018weakly}, WBG~\cite{nguyen2018weakly} as the state-of-the-art for weakly-supervised video recognition.
As no previous works have been carried on our setting, we need to implement and modify these works to fit our settings. Generally, we follow section~\ref{sec: baseline} to use untrimmed video works as the video feature extractor, then use FSL baselines to utilize the extracted features.
For the BL based methods, the model needs to use L$_2$ normalization for the output feature, which limits the use of works (e.g., \cite{lee2020uncertainty}) that rely on the feature norm.
For the MN based methods, the model is trained without the FC layer for classification, therefore methods rely on FC layers cannot be applied. For the classification-probability-based methods such as MIL~\cite{wang2017untrimmednets, lee2020uncertainty}, we simply average all segment features for the support set following TSN~\cite{wang2018temporal}, since no classification probability is available for the support set training samples.
To implement WBG~\cite{nguyen2019weakly}, which is also a background modeling based method, we also add a background class for both the BL and MN.
For action detection, we also implement the same set of baseline methods ourselves. Also, to compare with current works, we include the \cite{shou2017cdc,yang2018one,xu2020revisiting} which perform the few-shot untrimmed video detection. Note that these works utilize the full temporal annotations on base classes.
From Tab.~\ref{tab:sota_all} and Tab.~\ref{tab:sota_all_detection}, we can see that compared with the baselines implemented by us, we can achieve the best performance. Compared with current works with much more annotations, we can still achieve comparable performance or even higher.

\begin{figure}[t]
	\centering
	\includegraphics[width=0.8\columnwidth]{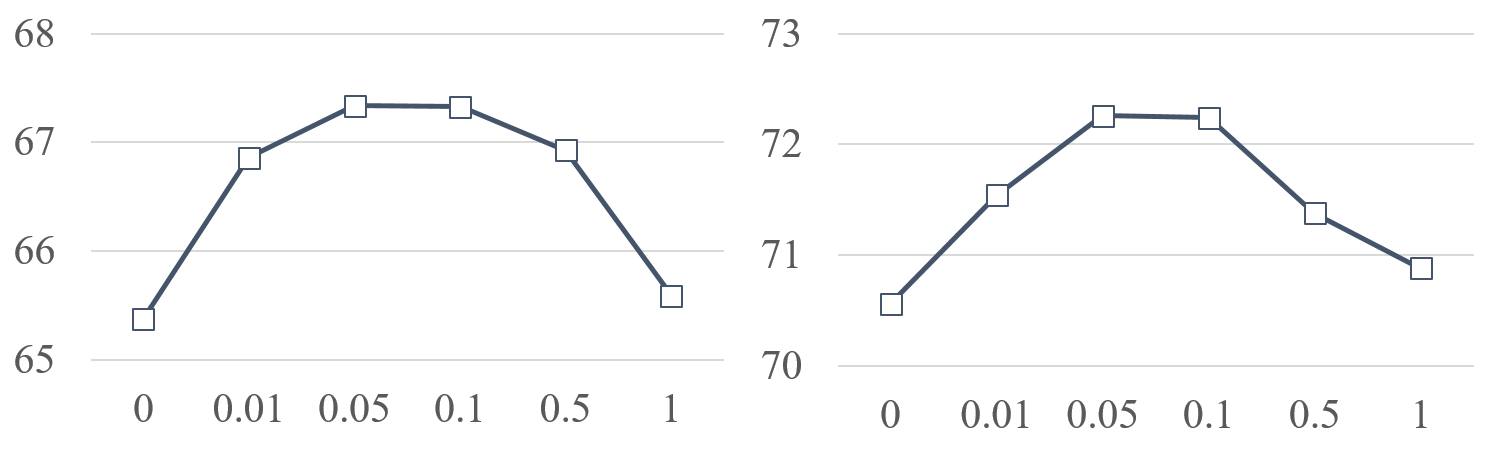}\vspace{-0.2cm}
	\caption{5-way 1-shot classification accuracy on ActivityNet v1.2 (left) or v1.3 (right) v.s. weight for $L_{contrast}$, i.e., $\gamma_1$.
	}\vspace{-0.3cm}
	\label{fig: contrastive_weight}
\end{figure}

\begin{figure}[t]
	\centering
	\includegraphics[width=0.8\columnwidth]{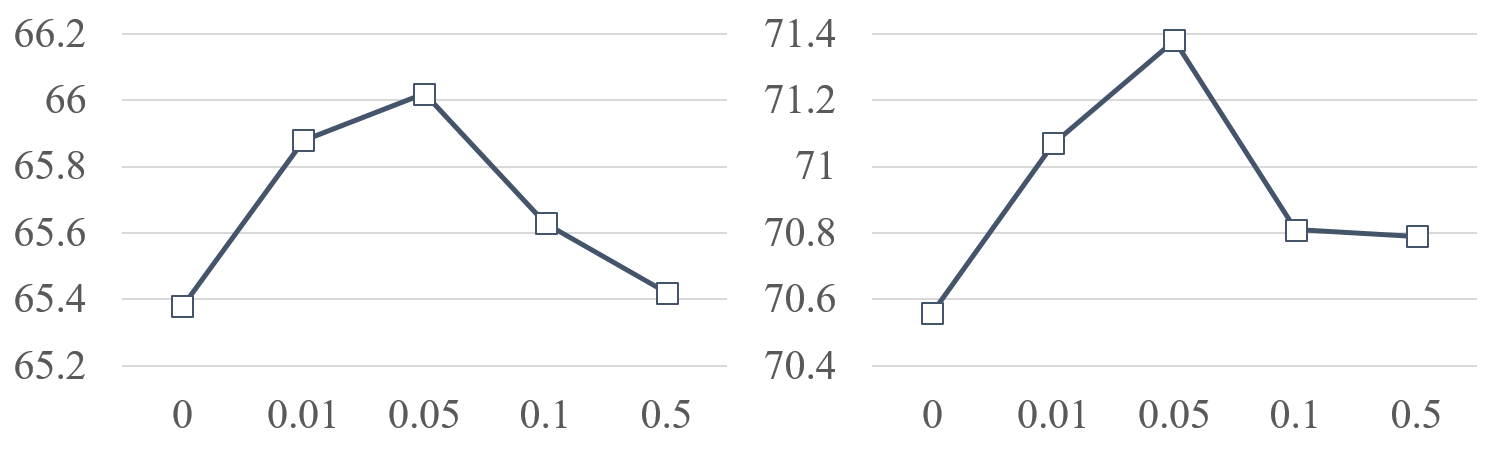}\vspace{-0.2cm}
	\caption{5-way 1-shot classification accuracy on ActivityNet v1.2 (left) or v1.3 (right) v.s. weight for $L_{bg-cls}$, i.e., $\gamma_2$.
	}\vspace{-0.4cm}
	\label{fig: bg_weight}
\end{figure}

\vspace{-0.2cm}
\subsection{Ablation study}

\subsubsection{Contribution of each module}

\begin{figure}
	\centering
	\includegraphics[width=0.7\columnwidth]{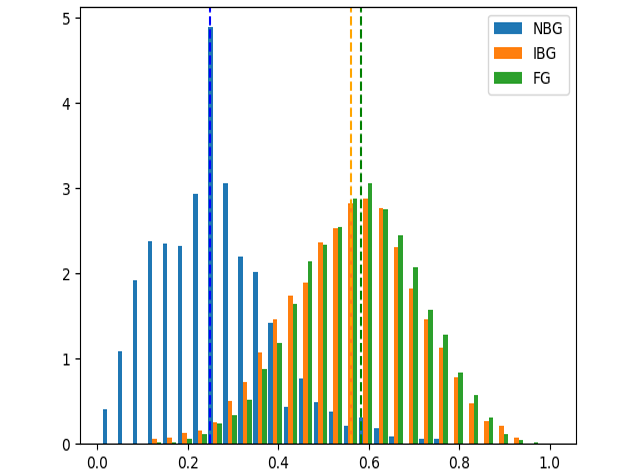} \vspace{-0.2cm}
	\caption{Distribution of the max logits of NBG, IBG and FG on our manually labeled dataset, where we can see a clear separation between NBG and IBG + FG. This annotation is NOT used in training.}
	\vspace{-0.3cm}
	\label{fig: max_logits}
\end{figure}

\begin{figure}[t]
	\centering
	\includegraphics[width=0.8\columnwidth,height=0.5\columnwidth]{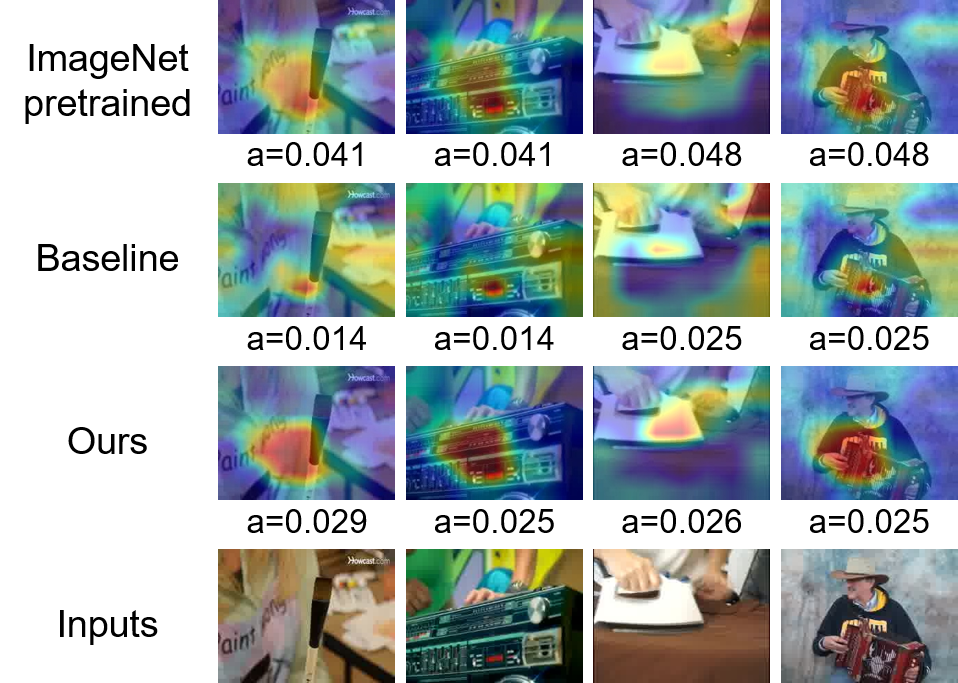}\vspace{-0.1cm}
	\caption{Heatmap on base-class BG with the activation ($a$) on each segment. ImageNet pretrained model is good at capturing objects. Compared with the first row, the baseline model does not capture informative objects in each BG segment, but the activation is lower. Compared with first two rows, our model can still capture informative objects in the IBG, with the activation lower than the first row, indicating the capability of distinguishing BG and FG.}\vspace{-0.3cm}
	\label{fig: heatmap}
\end{figure}

\begin{figure}[t]
	\centering
	\includegraphics[width=0.9\columnwidth]{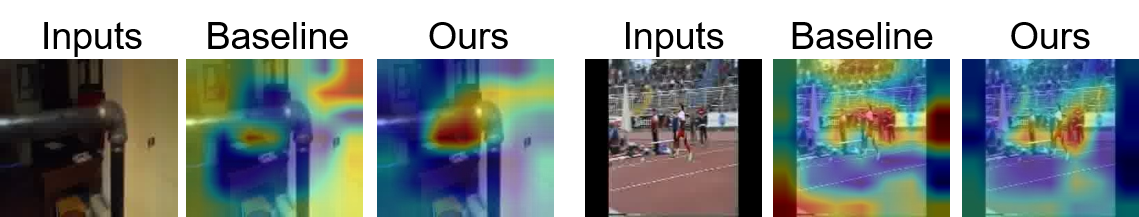}\vspace{-0.1cm}
	\caption{Heatmap on novel-class FG. Baseline model could not capture informative objects FG while our model could.}\vspace{-0.3cm}
	\label{fig: heatmap_novel}
\end{figure}

The contribution of each module is listed in Tab.~\ref{tab:ablation_all}. From this table, we can see every module has its own contribution to the performance.
Specifically,
as the pseudo-labeled segment is used in all modules, all improvements validate the correctness of the proposed pseudo-labeling method.
The self-weighted classification performs better than the soft-classification baseline, because it does not assume any global representation of the BG, showing better transferability across videos and classes.
The contrastive learning contributes the most because it helps the model to capture informative objects and movements by the self-supervision, and compressing the NBG class.

To verify the best choice of the weights for the proposed loss, we also include Fig.~\ref{fig: contrastive_weight} and Fig.~\ref{fig: bg_weight} to show the 5-way 1-shot classification accuracy against the corresponding loss weight. We can see that both weights peak at about 0.05.


\vspace{-0.1cm}
\subsubsection{Pseudo-labeling NBG, IBG and FG}
To verify our proposed criteria for pseudo-labeling in section~\ref{sec: finding} and \ref{sec: contrastive learning}, we manually labeled 177 untrimmed videos from base classes of ActivityNet v1.2 for its NBG and IBG. Together with its ground-truth FG, we plot the normalized base-class classification logits extracted by the baseline model on each video segment in Fig.~\ref{fig: max_logits}. The smaller the value in the x-axis, the closer the segment is from the uniform distribution; the higher the value in the y-axis, the more segments are in the corresponding bin, and the dashed vertical line denotes the mean value. From this figure we can see a clear separation between NBG and IBG + FG, while the IBG and FG are only marginally separated, therefore the max logits can indeed effectively identify the NBG and IBG + FG. Note that this annotation is \textbf{NOT} used during training.

\vspace{-0.1cm}
\subsubsection{Learning of BG and FG}

To study what the model has captured on base-class BG and novel-class FG, we visualize the corresponding heatmap in Fig.~\ref{fig: heatmap} and Fig.~\ref{fig: heatmap_novel}. In these figures, we averagely sample 25 segments from each video, and normalize each segments' activation by the sum of all segments' activation. Therefore, the average activation of each segment should be around 1/25=0.04. The activation of each segment is also plotted under each figure by $a$. From the base-class BG heatmap in Fig.~\ref{fig: heatmap}, we can see that the ImageNet pretrained model is good at capturing objects, but each segment's activation is near the average value 0.04, indicating it is not good at distinguishing BG and FG. Compared with the first row, the baseline model does not capture informative objects in video segments, but the activation is lower. Compared with first two rows, our model can still capture informative objects in IBG, with the activation lower than the first row, indicating the capability of distinguishing BG and FG. From the novel-class FG heatmap in Fig.~\ref{fig: heatmap_novel}, we can see that the baseline model could not capture informative objects in the novel-class FG while our model could.


\begin{figure}[t]
	\centering
	\includegraphics[width=1.0\columnwidth, height=0.43\columnwidth]{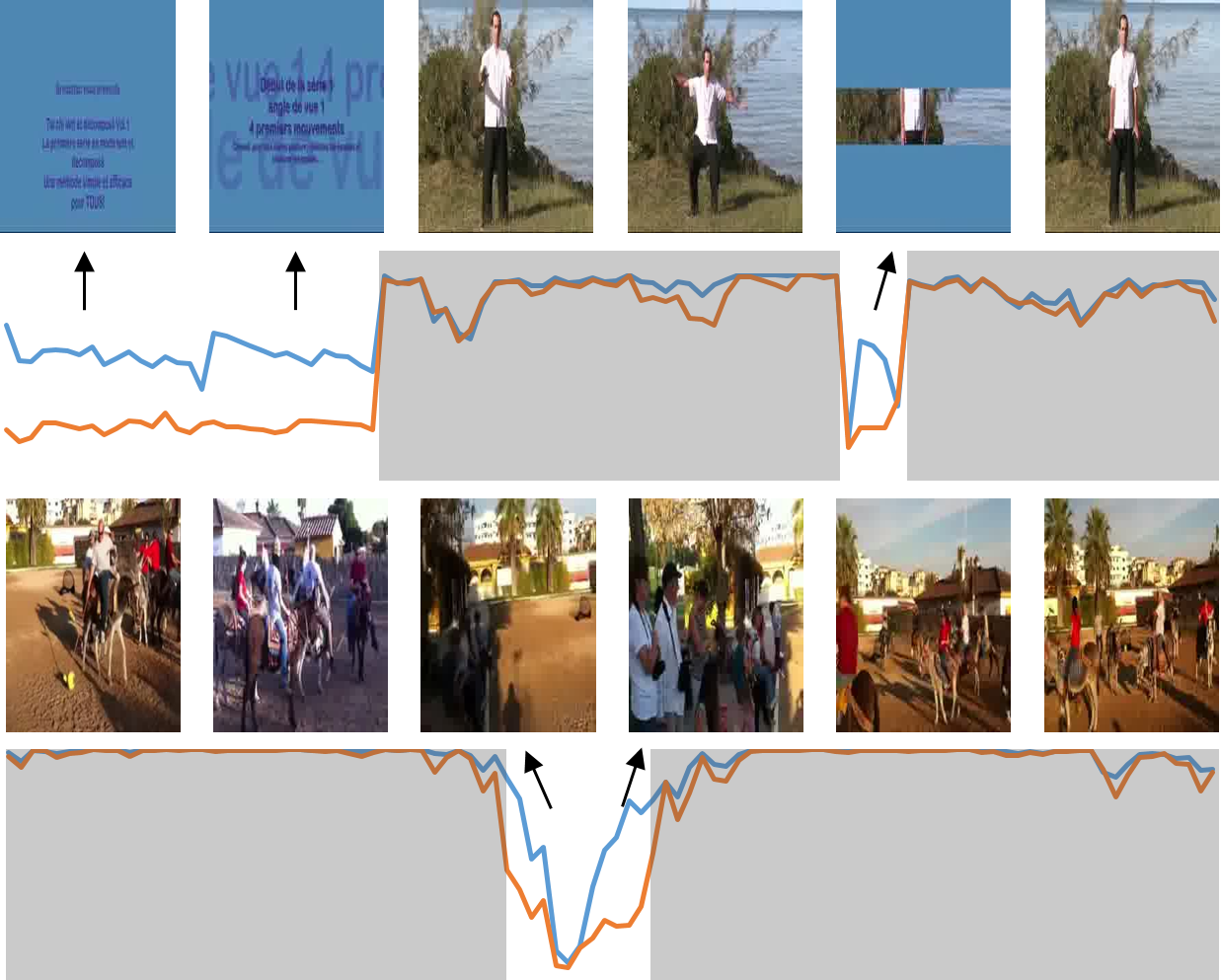}\vspace{-0.1cm}
	\caption{Qualitative evaluation of novel-class action detection. The shaded areas are the FG ground truth. The blue line denotes the action score predicted by the baseline model, while the orange line denotes that of our model. Our model shows better discriminability on the novel-class BG and FG.}\vspace{-0.4cm}
	\label{fig: detection}
\end{figure}

\vspace{-0.1cm}
\subsubsection{Qualitative evaluation of detection}

Besides the quantitative evaluation of action detection in Tab.~\ref{tab:sota_all_detection}, we also visualize the qualitative evaluation results in Fig.~\ref{fig: detection}. The shaded areas are the FG ground truth. The blue line denotes the action score predicted by the baseline model, while the orange line denotes that of our model which shows better discriminability on the novel-class BG and FG.

\vspace{-0.1cm}
\section{Conclusion}
To reduce the annotation of both the large amount of data and action locations, we proposed the Annotation-Efficient Video Recognition problem. To handle its challenges, we proposed (1) an open-set detection based method to find the NBG and FG, (2) a contrastive learning method for self-supervised learning of IBG and distinguishing NBG, and (3) a self-weighting mechanism for the better learning of IBG and FG.
Extensive experiments on ActivityNet v1.2 and v1.3 verified the effectiveness of the proposed methods.

\vspace{-0.15cm}
\section{Acknowledgments}
This work is partially supported by Key-Area Research and Development Program of Guangdong Province under contact No.2019B0101-53002, and grants from the National Natural Science Foundation of China under contract No. 61825101 and No. 62088102.

\bibliographystyle{ACM-Reference-Format}
\balance
\bibliography{zoilsen}

\end{document}